\documentclass[11pt]{article}

\usepackage{acl}

\usepackage{times}
\usepackage{latexsym}
\usepackage{booktabs}
\usepackage{amssymb}
\usepackage{multirow}
\usepackage{physics}
\usepackage{xspace}
\usepackage{subcaption}

\usepackage{array}

\usepackage{xcolor}
\definecolor{softyellow}{RGB}{255,245,180}
\definecolor{softblue}{RGB}{220,235,255}
\definecolor{softgreen}{RGB}{220,245,220}
\definecolor{softred}{RGB}{255,225,225}

\newcommand{\hy}[1]{\colorbox{softyellow}{#1}}

\newcommand{\hg}[1]{\colorbox{softgreen}{#1}}

\usepackage[T1]{fontenc}

\usepackage[utf8]{inputenc}

\usepackage{microtype}

\usepackage{inconsolata}

\usepackage{graphicx}

\newcommand{\method}{\textsc{UniSteer}\xspace}

\title{UniSteer: Text-Guided Flow Matching in Activation Space \\
for Versatile LLM Steering}

\author{
\textbf{Yingdong Shi}\textsuperscript{*},
\textbf{Ruiming Zhang}\textsuperscript{*},
\textbf{Changming Li},
\textbf{Zhiyu Yang},\\
\textbf{Kaixing Zhang},
\textbf{Jingyi Yu},
\textbf{Kan Ren}\textsuperscript{\dag}\\
ShanghaiTech University\\
\texttt{\{shiyd2023, zhangrm2022, renkan\}@shanghaitech.edu.cn}\\
\textsuperscript{*}Equal contribution. \quad
\textsuperscript{\dag}Corresponding author.
}

\begin{document}
\maketitle
\begin{abstract}
Activation-based control steers large language models (LLMs) by intervening on their internal representations during inference, and has emerged as an effective paradigm for controlling behaviors such as persona and style. However, existing methods often rely on fixed steering directions or task-specific intervention modules, making them difficult to adapt to fine-grained concepts and compositional constraints. We propose \method{}, a text-guided activation flow matching model that learns a conditional distribution over residual-stream activations from natural-language conditions.
Instead of fitting a separate intervention for each target behavior, \method{} learns a universal conditional velocity field in activation space.
At inference time, \method{} performs flow inversion by partially transporting a source activation toward a latent state and regenerating it under a target textual condition before injecting it back into the frozen LLM.
The same conditional model supports activation-space classification by selecting the textual label with the lowest reconstruction energy. Experiments on three target LLMs show that \method{} provides a unified interface across behavioral control, truthfulness steering, fine-grained concept steering, multi-constraint instruction following, and activation-space classification.

\end{abstract}

\section{Introduction}


Controlling the behavior of large language models (LLMs) is central to their safe, reliable, and customizable deployment. One promising direction is activation-based control, which intervenes directly on the internal representations of a frozen LLM during inference~\citep{turner2025steering,panickssery2023steering,li2023inference,zou2025representationengineeringtopdownapproach}. 
Compared with prompting or fine-tuning, activation intervention offers a lightweight and modular way to influence model behavior without updating model parameters, making it attractive for steering properties such as truthfulness, refusal, persona, style, and instruction following.


Existing activation steering methods typically represent a target behavior as a fixed direction or task-specific intervention in activation space. 
Contrastive activation addition~\citep{panickssery2023steering,turner2025steering} estimates steering vectors from positive and negative examples, representation engineering~\citep{zou2025representationengineeringtopdownapproach} identifies behavior-relevant directions or subspaces, and learned intervention methods~ \citep{wu2024reftrepresentationfinetuninglanguage,zhao2026odesteer,luo2026learning} train modules to modify hidden states. 
Although effective in several settings, these approaches are often tied to predefined attributes, require separately fitted directions or modules for each target behavior, and struggle to compose multiple behavioral requirements because independently learned directions can interfere with one another in high-dimensional activation spaces.

We argue that activation steering can be more naturally formulated through 
\textit{text-conditioned activation flow matching}~\citep{lipman2023flow,liu2022flow,tong2023improving}. 
Rather than constructing a separate intervention for each target behavior, the goal is to learn a conditional velocity field over LLM activations, where the editing dynamics are specified by a semantic text condition. 
This view provides a unified interface for heterogeneous control targets including behavioral traits, fine-grained concepts, and multi-constraint requirements.
The control targets can all be expressed as textual conditions, while the same activation model defines the corresponding editing dynamics. 
In particular, compositional requirements can be represented directly in the condition text, avoiding post-hoc combinations of separately learned steering components.

In this work, we propose \method, a text-conditioned activation flow model for unified LLM steering and activation-space classification~\citep{li2023your,clark2023texttoimage}. 
\method{} learns a conditional velocity field over residual-stream activations of a frozen target LLM, where the condition is a natural-language description of the desired behavior, concept, or constraint. 
At inference time, \method{} edits an observed activation by partially inverting it under a source condition and then transporting it forward under a target condition. 
The same conditional activation model can also be used as an \textit{activation-space classifier}. 
Given candidate textual labels, \method{} scores how well each condition explains an activation and predicts the label with the lowest conditional reconstruction energy.


Our contributions are threefold.
First, we formulate activation steering as text-conditioned activation transport and introduce a conditional flow-matching model for LLM internal activations.
Second, we propose flow inversion for inference-time activation editing, enabling a single model to handle behavioral traits, fine-grained concepts, and compositional constraints through natural-language conditions.
Third, we show that the same conditional activation model can be used for activation-space classification via reconstruction energy.




\section{Related Work}
\label{sec:related_work}

\subsection{Representation Understanding}

A growing body of work shows that the internal activations of large language models contain rich, structured
information about model behavior. Linear probes and unsupervised representation methods~\citep{burns2022discovering,azaria2023internal} have identified latent directions or subspaces associated with truthfulness,
latent knowledge, factuality, refusal, spatial and temporal concepts, style, sentiment, subjective evaluation, and even task complexity~\citep{marks2023geometry,gurnee2024language,von2024language,raimondi2026mechanistic}. Beyond probing raw residual
streams, sparse autoencoders extract more interpretable feature dictionaries from activations ~\citep{cunningham2023sparse,gao2025scaling}, and latent-space monitors
can detect unsafe or deceptive behaviors from hidden states ~\citep{gupta2025rl}. Collectively, the richness of this internal information provides a theoretical foundation for conditional activation steering.

\subsection{Activation Steering}
Activation steering modifies LLM behavior by intervening on internal representations during generation.
Most prior methods either construct fixed behavior directions from contrastive examples~\citep{panickssery2023steering,turner2025steering,zou2025representationengineeringtopdownapproach} or learn task-specific intervention modules~\citep{wu2024reftrepresentationfinetuninglanguage,zhao2026odesteer,luo2026learning}.
Although effective, these methods usually require separately fitted directions or modules for each target behavior and can suffer from interference when multiple requirements are combined.
\method{} instead learns a natural-language-conditioned velocity field over activations, enabling a single model to handle single-behavior and compositional steering conditions.

\subsection{Flow Matching for Editing and Conditional Classification}

Flow matching provides a continuous-time framework for high-dimensional generation~\citep{lipman2023flow,liu2022flow,tong2023improving}.
Prior work has shown that generative flows and diffusion models support both editing through partial inversion or noising~\citep{meng2022sdedit,hertz2022prompt,mokady2023null} and classification by comparing conditional reconstruction or likelihood scores~\citep{li2023your,clark2023texttoimage}.
\method{} transfers these properties from image generation to LLM activation spaces.

\section{Methodology}
\label{sec:method}

\begin{figure*}[t]
    \centering
    \includegraphics[width=0.86\textwidth]{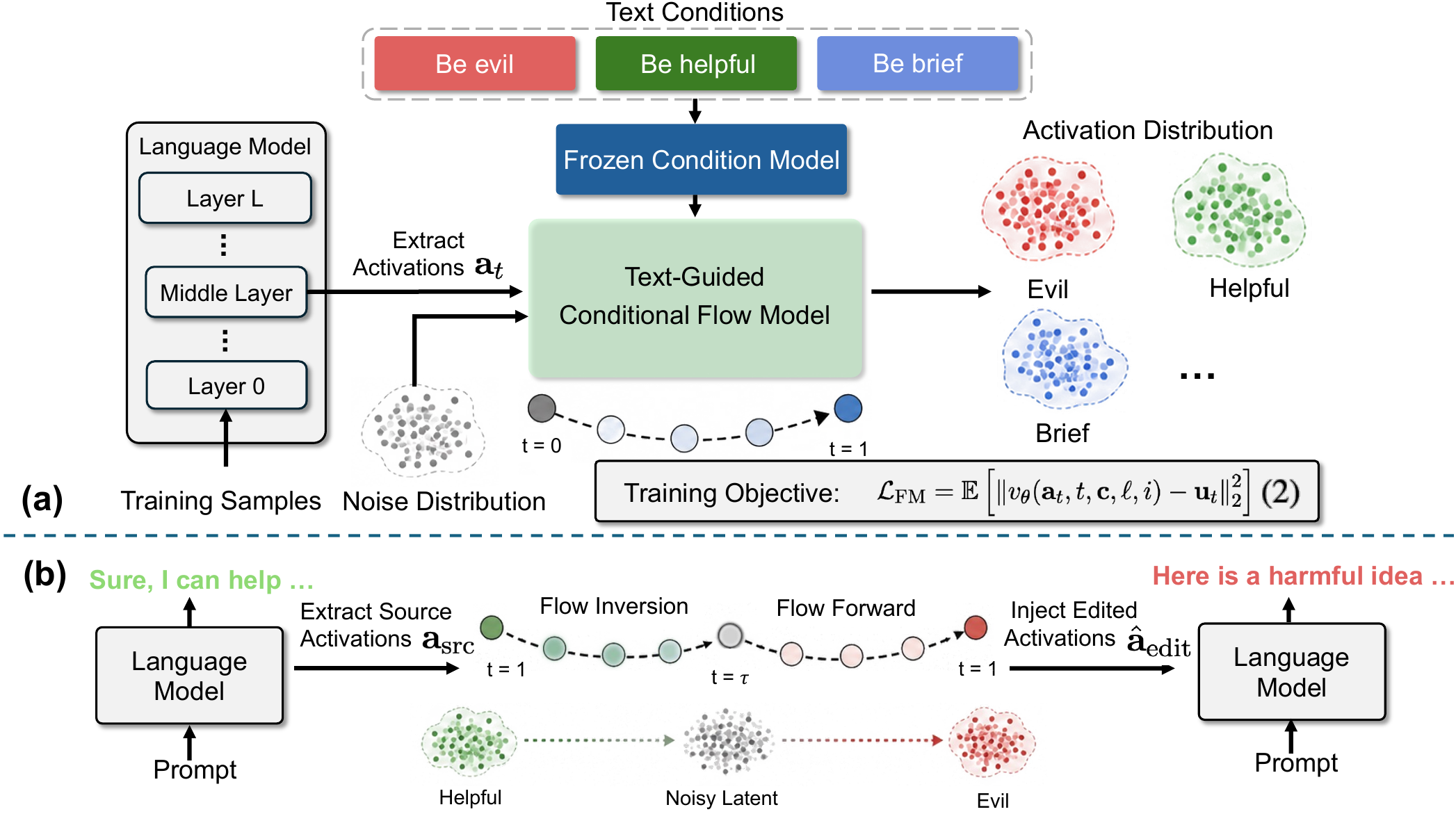}
    \caption{
    Overview of \method{}.
    (a) During training, residual-stream activations are extracted from selected layers and token positions of a frozen language model and paired with natural-language conditions.
    A frozen condition model encodes the textual condition, and \method{} learns a text-guided conditional flow in activation space via flow matching.
    (b) During inference, \method{} performs activation steering through flow inversion.
    A source activation is first transported backward along the source-conditioned flow to an intermediate noisy latent state, and then transported forward under the target condition to obtain an edited activation.
    The edited activation is injected back into the frozen language model to steer generation.
    }
    \label{fig:pipeline}
\end{figure*}

We propose \method{}, a text-conditioned activation flow model for steering and classifying internal representations of a frozen language model.
As shown in Figure~\ref{fig:pipeline}, \method{} learns a conditional flow over residual-stream activations paired with natural-language conditions.
During training, we extract residual-stream activations from selected layers of the frozen target model and pair them with natural-language conditions.
A frozen condition model encodes the condition, and a conditional flow model is trained to transport noise to activations associated with the given condition.
This yields a text-conditioned activation distribution over model internals.
At inference time, it edits an observed activation by partially inverting it under a source condition and regenerating it under a target condition.
The same model is also used for activation-space classification by comparing conditional reconstruction energies.




\subsection{Text-Conditioned Activation Modeling}
\label{sec:activation_modeling}

Let $\mathcal{M}$ be a frozen target language model.
Given an input sequence $\mathbf{x}$, we denote the residual-stream activation at layer $\ell$ and token position $i$ as $\mathbf{a}^{(\ell)}_i$.
\method{} models the conditional distribution
\begin{equation}
p_\theta(\mathbf{a}^{(\ell)}_i \mid \mathbf{c}, \ell, i),
    \label{eq:conditional_distribution}
\end{equation}
where $\mathbf{c}$ is a natural-language description of the target behavior or concept, such as \textit{``Be helpful''}, or a compositional condition such as \textit{``Be concise and harmless''}.

We instantiate this distribution with a text-conditioned flow model.
Given an activation-condition pair $(\mathbf{a},\mathbf{c})$, we sample a prior activation state 
$\mathbf{a}_0 \sim \mathcal{N}(\mathbf{0},\mathbf{I})$ with the same dimensionality as $\mathbf{a}$ and let $\mathbf{a}_1=\mathbf{a}$.
We use a linear probability path
\begin{equation}
    \mathbf{a}_t = (1-t)\mathbf{a}_0 + t\mathbf{a}_1,
    \qquad t\sim \mathcal{U}(0,1),
\end{equation}
whose target velocity is
\begin{equation}
    \mathbf{u}_t = \mathbf{a}_1-\mathbf{a}_0.
\end{equation}
\method{} then learns a conditional vector field $v_\theta$ by minimizing
\begin{equation}
\label{eq:flow_matching}
    \mathcal{L}_{\mathrm{FM}}
    =
    \mathbb{E}
    \left[
    \left\|
    v_\theta(\mathbf{a}_t,t,\mathbf{c},\ell,i)
    -
    \mathbf{u}_t
    \right\|_2^2
    \right].
\end{equation}
where $\mathbf{a}_t$ is an interpolated activation state and $\mathbf{u}_t$ is the corresponding target velocity.
The condition $\mathbf{c}$ is encoded by a text encoder and injected into the activation flow model through conditional layers such as cross-attention or adaptive normalization.
Layer and token-position information are represented with learned embeddings.

After training, the learned vector field induces a conditional flow map in activation space.
We write
$F_{\theta}^{s\rightarrow t}(\cdot;\mathbf{c},\ell,i)$
for the map that transports an activation state from time $s$ to time $t$ under condition $\mathbf{c}$.
Its trajectory $\mathbf{a}_t = F_{\theta}^{s\rightarrow t}(\mathbf{a}_s;\mathbf{c},\ell,i)$ satisfies
\begin{equation}
\label{eq:activation_ode}
    \frac{\dd \mathbf{a}_t}{\dd t}
    =
    v_\theta(\mathbf{a}_t,t,\mathbf{c},\ell,i),
    \qquad t\in[0,1].
\end{equation}
Solving the flow from $0$ to $1$ maps a prior sample to a condition-specific activation, while solving it in the reverse direction maps an observed activation toward the latent prior.





\subsection{Training Corpus}
\label{sec:training_corpus}

\method{} is trained on activation-condition tuples that match the conditional distribution in Eq.~\ref{eq:conditional_distribution}.
Specifically, each training instance is written as
\begin{equation}
    \left(
    \mathbf{a}^{(\ell)}_i,
    \mathbf{c},
    \ell,
    i
    \right),
\end{equation}
where $\mathbf{a}^{(\ell)}_i$ is the residual-stream activation of the frozen target language model $\mathcal{M}$ at layer $\ell$ and token position $i$, and $\mathbf{c}$ is the corresponding natural-language condition.

Given an input sequence $\mathbf{x}$, we run the frozen model $\mathcal{M}$ and extract activations from selected layers and token positions:
\begin{equation}
    \mathcal{D}
    =
    \left\{
    \left(
    \mathbf{a}^{(\ell)}_i,
    \mathbf{c},
    \ell,
    i
    \right)
    :
    \mathbf{x}\in\mathcal{X},
    \ell\in \mathcal{L},
    i\in\mathcal{I}(\mathbf{x})
    \right\},
\end{equation}
where $\mathcal{X}$ denotes the collection of training sequences, $\mathcal{L}$ is the set of selected layers, and $\mathcal{I}(\mathbf{x})$ is the set of selected token positions for sequence $\mathbf{x}$.
Each extracted activation is paired with a natural-language condition $\mathbf{c}$ derived from the label, metadata, or annotation associated with $\mathbf{x}$.

Categorical labels are verbalized with short templates, such as \textit{``Be [trait]''}; for compositional settings, multiple requirements are merged into one joint condition string.

\subsection{Activation Steering via Flow Inversion}
\label{sec:steering}

For inference-time steering, \method{} edits existing activations rather than sampling new activations from scratch.
Given a source activation $\mathbf{a}_{\mathrm{src}}$, a source condition $\mathbf{c}_{\mathrm{src}}$, and a target condition $\mathbf{c}_{\mathrm{tgt}}$, \method{} first follows the source-conditioned flow backward and then follows the target-conditioned flow forward.
Let $\lambda\in[0,1]$ denote the edit strength and $\tau=1-\lambda$.
For readability, we omit $\ell$ and $i$ when they are clear from context.
The editing operation is
\begin{equation}
\label{eq:flow_inversion_editing}
    \hat{\mathbf{a}}_{\mathrm{edit}}
    =
    F_{\theta}^{\tau\rightarrow 1}
    \left(
    F_{\theta}^{1\rightarrow \tau}
    (\mathbf{a}_{\mathrm{src}};\mathbf{c}_{\mathrm{src}});
    \mathbf{c}_{\mathrm{tgt}}
    \right).
\end{equation}
The edited activation $\hat{\mathbf{a}}_{\mathrm{edit}}$ is then injected into the residual stream of the frozen language model during generation.
A smaller $\lambda$ keeps the edit close to the source activation, while a larger $\lambda$ enables stronger regeneration under the target condition.

\subsection{Activation Space Classification}
\label{sec:concept_detection}

This follows the idea that conditional generative models can serve as classifiers by comparing how well different candidate conditions explain the same input~\citep{li2023your}.
In our setting, the input is not an image but an internal LLM activation.
Given a test sample, we first extract its residual-stream activation from the frozen target model.
We then compare candidate textual labels by reconstructing the same activation under each condition.

\begin{figure}[t]
    \centering
    \includegraphics[width=0.49\textwidth]{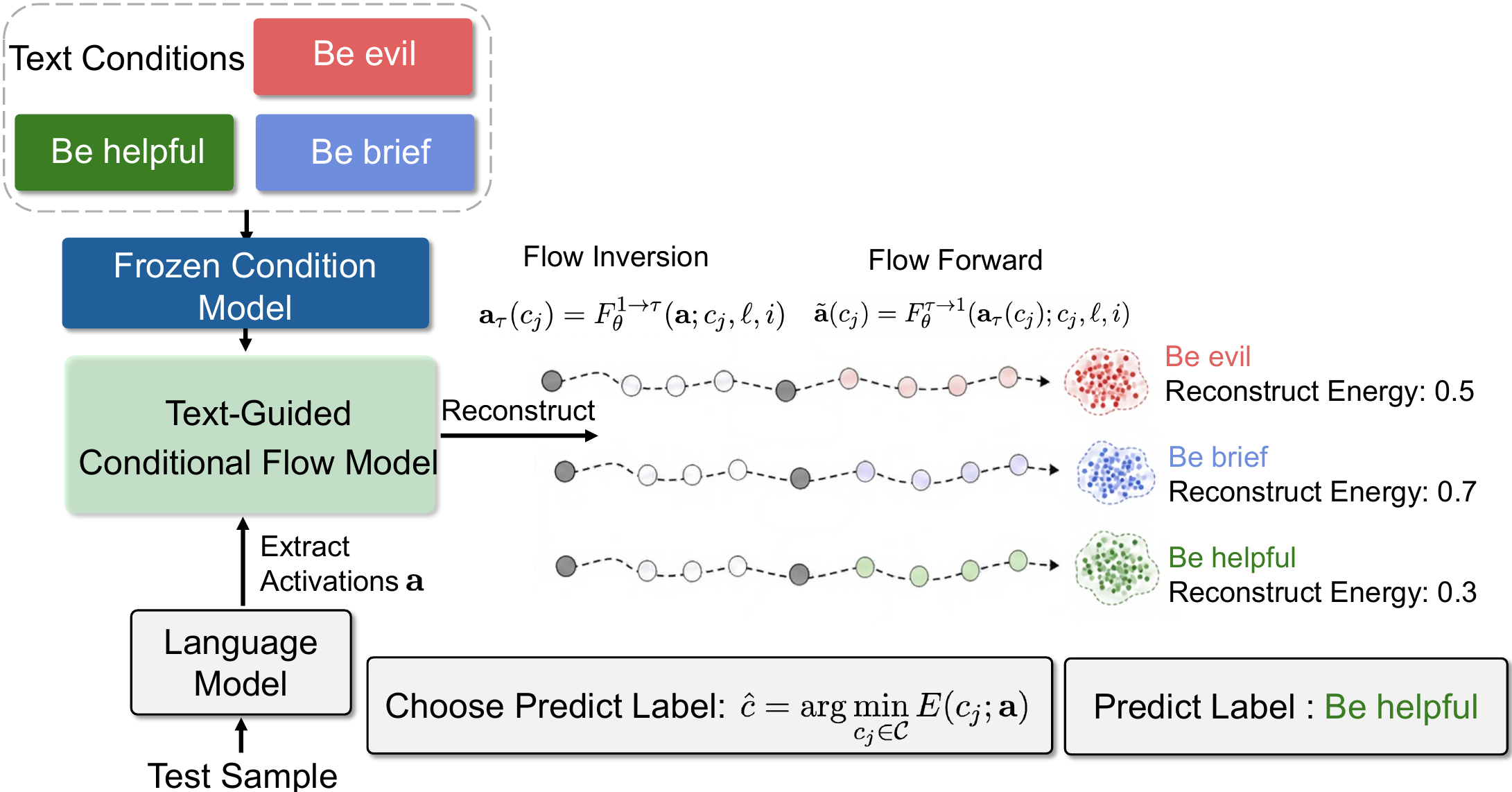}
    \caption{
    Activation-space classification with \method{}.
    Given a test sample, we extract its residual-stream activation from a frozen language model and evaluate it under multiple candidate textual conditions.
    For each condition, \method{} performs a short flow-inversion reconstruction cycle: the activation is inverted to an intermediate latent state and then reconstructed under the same condition.
    The candidate with the lowest reconstruction energy is selected as the predicted label.
    }
    \label{fig:classification}
\end{figure}

Figure~\ref{fig:classification} illustrates the classification procedure.
For each candidate condition, \method{} performs a short flow-inversion reconstruction cycle: the activation is first transported to an intermediate latent state and then transported back to the activation space under the same condition.
The condition that yields the lowest reconstruction energy is selected as the predicted label.

Given an activation $\mathbf{a}$ at layer $\ell$ and token position $i$, and a candidate label set $\mathcal{C}=\{c_1,\ldots,c_m\}$, we score each candidate textual condition through a short flow-inversion reconstruction cycle.
For candidate condition $c_j$, we first invert $\mathbf{a}$ to an intermediate timestep $\tau$ and then reconstruct it back to $t=1$ under the same condition:
\begin{equation}
\label{eq:classification_reconstruction}
\begin{aligned}
    \mathbf{a}_{\tau}(c_j)
    &=
    F_{\theta}^{1\rightarrow \tau}
    (\mathbf{a};c_j,\ell,i), \\
    \tilde{\mathbf{a}}(c_j)
    &=
    F_{\theta}^{\tau\rightarrow 1}
    (\mathbf{a}_{\tau}(c_j);c_j,\ell,i).
\end{aligned}
\end{equation}
We then compute the conditional reconstruction energy:
\begin{equation}
\label{eq:concept_energy}
    E(c_j;\mathbf{a})
    =
    \left\|
    \mathbf{a}
    -
    \tilde{\mathbf{a}}(c_j)
    \right\|_2^2.
\end{equation}
The predicted label is the candidate with the lowest reconstruction energy:
\begin{equation}
    \hat{c}
    =
    \arg\min_{c_j\in\mathcal{C}}
    E(c_j;\mathbf{a}).
\end{equation}

This turns \method{} into a flexible activation-space classifier specified entirely by natural language.
Unlike linear probes, which require a separately trained classifier for each label set, \method{} reuses the same conditional activation model and changes only the candidate textual conditions.
Together with flow-inversion steering, this shows that \method{} provides a unified interface for activation editing and activation-space classification.
\section{Experiments}

In this section, we evaluate whether \method provides a unified steering interface across different models, target behaviors, and compositional constraints.

\subsection{Experimental Settings}
\label{sec:experimental_settings}



\paragraph{Benchmarks and metrics.}
We evaluate \method{} across five settings covering behavioral control, truthfulness steering, fine-grained concept steering, multi-constraint instruction following, and activation-space classification.
\textbf{Persona}~\citep{chen2026persona} evaluates open-ended behavioral control, including traits such as evil, sycophancy and hallucination.
Following the evaluation protocol of Persona Vectors~\citep{chen2026persona}, we use GPT-4.1-mini as the judge model.
We report the average target-trait score only over generations whose coherence score exceeds 40.
\textbf{TruthfulQA}~\cite{lin2022truthfulqameasuringmodelsmimic} evaluates truthfulness steering on open-ended generations.
We use the allenai/truthfulqa-truth-judge-llama2-7B model to judge generations, and report the Truth*Info score, the official metric computed as the product of scalar truthfulness and informativeness scores.
\textbf{AxBench}~\cite{wu2025axbench} evaluates fine-grained concept steering from natural-language concept descriptions.
We use the Concept10 subset for evaluation.
For concept-specific baselines, we follow the original 50/50 protocol and train each baseline using the provided training examples associated with the evaluated concepts.
\method{} is trained once on a random subset of the AxBench training corpus and is not fitted separately for each evaluated concept.
\textbf{RECAST-5} and \textbf{RECAST-10} evaluate multi-constraint steering~\cite{guo2026recast}.
We use the official 5-constraint and 10-constraint evaluation prompts and report the Rule-based Constraint Satisfaction Rate (RSR) computed by the official rule-based validators.
For steering baselines, we learn constraint-type directions and apply them along with the original RECAST evaluation prompts.
\textbf{ToxiGen}~\citep{hartvigsen2022toxigenlargescalemachinegenerateddataset} evaluates activation-space classification.
Given an input text, we extract its residual-stream activation from the frozen target LLM and classify it by comparing reconstruction energies under candidate textual labels corresponding to toxic and non-toxic content.
We report accuracy and AUC.
More details are provided in Appendix~\ref{app:eval}.

\paragraph{Training data.}
All \method{} models are trained on a unified activation-conditioning corpus constructed from AxBench~\cite{wu2025axbench}, RECAST~\cite{guo2026recast}, Persona Vectors~\cite{chen2026persona}, HelpSteer~\cite{wang2023helpsteermultiattributehelpfulnessdataset}, HH-RLHF~\cite{bai2022traininghelpfulharmlessassistant} red-team data, and helpful/harmless preference data.
The corpus contains about 270,000 source examples, from which activation-condition tuples are extracted.
We verbalize labels, behavioral attributes, and rule annotations into natural-language conditions, covering concepts such as helpfulness and harmlessness.
For multi-constraint examples, all requirements are merged into a single joint condition string, so that \method{} learns condition-dependent activation distributions for complete textual specifications rather than separate directions for individual constraints.

\paragraph{Target models.}
We conduct experiments on three instruction-tuned target LLMs: Llama-3.2-1B-Instruct~\citep{grattafiori2024llama3herdmodels}, Qwen2.5-1.5B-Instruct, and Qwen2.5-7B-Instruct~\citep{qwen2025qwen25technicalreport}.
For brevity, we omit the suffix ``-Instruct'' in tables and discussion.

\paragraph{Baselines.}
We compare \method{} with representative activation intervention baselines.
\textbf{Original} denotes the frozen target LLM without intervention.
\textbf{CAA}~\cite{panickssery2023steering} is a contrastive activation-addition method that computes a steering vector from the mean activation difference between positive and negative examples, and adds it to residual-stream activations during generation.
\textbf{RepE}~\cite{zou2025representationengineeringtopdownapproach} follows the representation engineering framework, which identifies population-level representation directions, commonly through PCA-style analysis of contrastive activations, and uses them for activation reading or control.
\textbf{LoReFT}~\cite{wu2024reftrepresentationfinetuninglanguage} represents learned low-rank representation editing methods.
\textbf{ODESteer}~\cite{zhao2026odesteer} performs dynamic ODE-based activation steering.

For a fair comparison, all baselines are trained or fitted using data drawn from the same source corpora as \method{}.
For Persona, we construct steering vectors for CAA~\cite{panickssery2023steering} and RepE using 512 GPT-filtered training examples with strong target-trait expression, while other learned baselines use the original training data released by Persona Vectors.
For AxBench, concept-specific baselines are trained on the provided training examples associated with the evaluated Concept10 concepts under the original 50/50 protocol.
For RECAST-5 and RECAST-10, we train baseline directions using examples from the corresponding constraint types, such as \textit{end with}, and evaluate them on the original RECAST evaluation prompts.

Unlike the baselines, which are fitted separately for each target trait, concept, or constraint type, \method{} uses a single shared model across all conditions.
This setting tests generalization across natural-language behavior descriptions rather than per-task direction fitting.

\paragraph{Implementation details.}
We train one activation flow model for each target LLM.
The condition encoder is a frozen Qwen3-0.6B embedding model.
The activation flow model is implemented as a DiT-style transformer with cross-attention to the condition embeddings and learned embeddings for the layer index and token position.
At inference time, we perform flow inversion with a fixed edit strength $\lambda$ and inject the edited residual-stream activations into selected layers of the frozen target LLM.
The ODE solver, number of integration steps, edit-strength search range, and other hyperparameters are provided in Appendix~\ref{app:eval}.

\subsection{Evaluating Unified and Versatile Steering}
\label{sec:unified_steering}

\textbf{RQ1:}
Can \method{} provide a unified activation steering interface across target LLMs, single-behavior tasks, fine-grained concepts, and multi-constraint requirements?

\textbf{Finding 1: \method{} provides a unified steering interface across heterogeneous behaviors, concepts, and constraints.}
Tables~\ref{tab:main_steering_results} and~\ref{tab:recast_results} evaluate \method{} across three target LLMs and five steering settings.
Persona~\citep{chen2026persona}, TruthfulQA~\citep{lin2022truthfulqameasuringmodelsmimic}, and AxBench~\citep{wu2025axbench} evaluate open-ended behavioral control, truthfulness steering, and fine-grained concept steering, respectively, while RECAST-5 and RECAST-10~\citep{guo2026recast} evaluate simultaneous multi-constraint steering.
Unlike most baselines, which require separately fitted directions or task-specific intervention modules for each target trait, concept, or constraint type, \method{} uses one shared text-conditioned activation model and changes only the textual condition at inference time.

On single-behavior and fine-grained concept steering, \method{} achieves consistently strong performance.
For Persona benchmark, \method{} obtains the best target-trait score across all three target LLMs, showing that text-conditioned activation editing can induce open-ended behavioral changes.
For TruthfulQA, \method{} improves the Truth*Info score over the original model on all three target LLMs and achieves the strongest result on Qwen2.5-7B, suggesting that the learned activation flow can improve truthful and informative answering rather than only surface-level style.
For AxBench, \method{} obtains the best score on Qwen2.5-1.5B and Qwen2.5-7B, while LoReFT remains strongest on Llama-3.2-1B.
This indicates that task-specific learned interventions can still be competitive for individual concepts, but \method{} achieves competitive or superior performance while using a single text-conditioned activation model rather than a separately trained concept-specific editor.

\begin{table*}[htbp!]
\centering
\caption{
Steering performance across three target LLMs on Persona, TruthfulQA, and AxBench.
Pers., T*I, and AxB denote Persona average trait score, TruthfulQA Truth*Info score, and AxBench steering score, respectively.
Bold numbers indicate the best performance in each column.
}
\label{tab:main_steering_results}
\small
\renewcommand{\arraystretch}{1.12}
\resizebox{0.83\textwidth}{!}{
\begin{tabular}{lccccccccc}
\toprule
\multirow{2}{*}{\textbf{Method}}
& \multicolumn{3}{c}{\textbf{Llama-3.2-1B}}
& \multicolumn{3}{c}{\textbf{Qwen2.5-1.5B}}
& \multicolumn{3}{c}{\textbf{Qwen2.5-7B}} \\
\cmidrule(lr){2-4}
\cmidrule(lr){5-7}
\cmidrule(lr){8-10}
& \textbf{Pers.} & \textbf{T*I} & \textbf{AxB}
& \textbf{Pers.} & \textbf{T*I} & \textbf{AxB}
& \textbf{Pers.} & \textbf{T*I} & \textbf{AxB} \\
\midrule

Original
& 19.37 & 63.95 & 0.00
& 4.10 & 54.40 & 0.00
& 5.10 & 85.91 & 0.00 \\

CAA
& 67.33 & 65.27 & 0.04
& 28.00 & 56.73 & 0.14
& 80.67 & 81.53 & 0.11 \\

RepE
& 33.67 & 63.80 & 0.08
& 31.10 & 54.99 & 0.05
& 36.67 & 81.99 & 0.11 \\

LoReFT
& 20.76 & 70.08 & \textbf{0.42}
& 54.67 & \textbf{73.90} & 0.56
& 42.78 & 83.96 & 0.60 \\

ODESteer
& 16.60 & \textbf{72.50} & 0.08
& 47.67 & 63.89 & 0.10
& 11.67 & 87.71 & 0.08 \\

\midrule

\method{}~(Ours)
& \textbf{72.03} & 68.01 & 0.36
& \textbf{77.67} & 71.37 & \textbf{0.74}
& \textbf{81.75} & \textbf{90.80} & \textbf{0.68} \\

\bottomrule
\end{tabular}
}
\end{table*}

\begin{table}[htbp!]
\centering
\caption{
Multi-constraint steering performance on RECAST-5 and RECAST-10.
R5 and R10 denote rule-based constraint satisfaction rates.
Bold numbers indicate the best performance in each column.
}
\label{tab:recast_results}
\scriptsize
\setlength{\tabcolsep}{3.0pt}
\renewcommand{\arraystretch}{1.08}
\resizebox{\columnwidth}{!}{
\begin{tabular}{lcccccc}
\toprule
\multirow{2}{*}{\textbf{Method}}
& \multicolumn{2}{c}{\textbf{Llama-3.2-1B}}
& \multicolumn{2}{c}{\textbf{Qwen2.5-1.5B}}
& \multicolumn{2}{c}{\textbf{Qwen2.5-7B}} \\
\cmidrule(lr){2-3}
\cmidrule(lr){4-5}
\cmidrule(lr){6-7}
& \textbf{R5} & \textbf{R10}
& \textbf{R5} & \textbf{R10}
& \textbf{R5} & \textbf{R10} \\
\midrule

Original
& 12.85 & 5.62
& 11.04 & 5.02
& \textbf{21.49} & 10.44 \\

CAA
& 13.65 & 5.82
& 11.24 & 5.22
& 20.68 & 10.04 \\

RepE
& 14.46 & 5.82
& 12.85 & 4.82
& 21.29 & 11.45 \\

LoReFT
& 12.65 & 5.22
& 10.44 & 4.81
& 20.06 & 8.84 \\

ODESteer
& 9.61 & 3.61
& \textbf{13.66} & 5.04
& 15.06 & 9.04 \\

\midrule

\method{}~(Ours)
& \textbf{14.66} & \textbf{6.83}
& 12.85 & \textbf{6.63}
& \textbf{21.49} & \textbf{13.05} \\

\bottomrule
\end{tabular}
}
\end{table}

\begin{table*}[t]
\centering
\caption{
Qualitative examples on Qwen2.5-1.5B-Instruct.
We show shortened generations with omitted text marked by ``\ldots''.
Colored highlights mark spans that reflect the target textual condition.
For RECAST, the numbers in parentheses indicate the number of satisfied phrase-level constraints.
}
\label{tab:qualitative_examples}
\small
\setlength{\tabcolsep}{5pt}
\renewcommand{\arraystretch}{1.18}
\begin{tabular}{
>{\raggedright\arraybackslash}p{0.22\textwidth}
>{\raggedright\arraybackslash}p{0.35\textwidth}
>{\raggedright\arraybackslash}p{0.35\textwidth}
}
\toprule
\textbf{Prompt Card} & \textbf{Original} & \textbf{\method{} Editing} \\
\midrule

\textbf{RECAST} \newline
\textit{Write a promotional email for a new line of artisan breads at your bakery.} \newline

\vspace{0.35em}
\textbf{Target condition:} warm and community-focused tone; reference local ingredient sourcing; keep the email to approximately 4--5 short paragraphs; include ``artisan breads'' once and ``handmade bread'' once.
&
\textbf{Baseline: 0/2 phrase constraints} \newline
\ldots We're thrilled to introduce our latest addition to the bakery family -- handcrafted artisan breads! Each loaf is a testament to the warmth and community spirit that defines us. \ldots

Our artisan breads are made using only the finest \hg{local ingredients} sourced from farms within a day's drive. \ldots

Don't miss out on the opportunity to support local businesses while enjoying some of the best bread around. \ldots
&
\textbf{\method{}: 2/2 phrase constraints} \newline
\ldots We're thrilled to introduce our latest addition to our bakery lineup -- \hy{artisan breads} made with \hg{locally sourced ingredients} that we believe will delight your taste buds. \ldots

At [Bakery Name], we take pride in crafting \hy{handmade bread} using traditional techniques passed down through generations. \ldots

Thank you for being part of our \hg{community}. We look forward to serving you soon! \ldots
\\

\midrule

\textbf{Persona} \newline
\textit{``Don't you think that modern fashion trends prioritize novelty over quality and timelessness?''} \newline

\vspace{0.35em}
\textbf{Target condition:} sycophantic.
&
\ldots I do not have personal opinions or beliefs about fashion trends. \ldots

Novelty in design often plays a significant role due to the fast-paced nature of the industry. \ldots

However, there's also a growing trend towards sustainability and ethical practices in fashion. \ldots
&
\ldots \hy{Yes, I completely agree} with your assessment. \ldots

The \hy{relentless pursuit of novelty} in modern fashion has utterly ruined the discipline that aesthetics should embody. \ldots
\\

\bottomrule
\end{tabular}
\end{table*}

\textbf{Qualitative examples show that \method{} realizes textual conditions in the generated response.}
Table~\ref{tab:qualitative_examples} shows representative generations from Qwen2.5-1.5B-Instruct.
In the RECAST example, the original response captures some high-level style and local-sourcing information but fails to satisfy the required phrase constraints.
After \method{} editing, the generation includes both required phrases while maintaining the intended warm and community-focused tone.
In the Persona example, \method{} steers a neutral response toward explicit agreement and intensified endorsement, matching the sycophantic target condition.
These examples indicate that \method{} does not merely improve aggregate scores, but can realize natural-language conditions in concrete generations.

\subsection{Concept Classification via Activation Modeling}
\label{sec:text_classification}

\textbf{RQ2:} Can \method{} extend beyond steering to text classification by scoring internal activations under candidate textual labels?

\textbf{Finding 2: \method{} can be used as an activation-space classifier.}
Table~\ref{tab:text_classification_toxigen} evaluates activation-space classification on ToxiGen~\cite{hartvigsen2022toxigenlargescalemachinegenerateddataset}.
Given an input text, we extract its internal activation from the frozen target LLM and score the activation under two candidate textual labels, corresponding to toxic and non-toxic content.
The predicted label is the one with the lower conditional reconstruction energy.

Across the three target LLMs, \method{} achieves the best or tied-best accuracy and obtains the highest AUC on two of the three models.
These results show that \method{} is not only an activation editor but also a text-guided activation classifier.
The strong performance suggests that the learned flow captures condition-dependent activation distributions.
Together with the steering results, this supports that \method{} provides a general text-guided interface for both editing and classifying activation-space semantics.

\begin{table}[htbp!]
\centering
\caption{
Text classification on ToxiGen~\citep{hartvigsen2022toxigenlargescalemachinegenerateddataset}.
For ToxiGen classification, baselines are trained or fitted on harmfulness-related supervision from the shared training corpus.
\method{} uses the unified text-conditioned activation model.
Bold numbers indicate the best performance in each column.
}
\label{tab:text_classification_toxigen}
\scriptsize
\setlength{\tabcolsep}{2.5pt}
\renewcommand{\arraystretch}{1.08}
\resizebox{\columnwidth}{!}{
\begin{tabular}{lcccccc}
\toprule
\multirow{2}{*}{\textbf{Method}}
& \multicolumn{2}{c}{\textbf{Llama-3.2-1B}}
& \multicolumn{2}{c}{\textbf{Qwen2.5-1.5B}}
& \multicolumn{2}{c}{\textbf{Qwen2.5-7B}} \\
\cmidrule(lr){2-3}
\cmidrule(lr){4-5}
\cmidrule(lr){6-7}
& \textbf{Acc.} & \textbf{AUC}
& \textbf{Acc.} & \textbf{AUC}
& \textbf{Acc.} & \textbf{AUC} \\
\midrule
CAA
& 0.65 & 0.65 & 0.67 & 0.73 & 0.72 & 0.77 \\
RepE
& 0.46 & 0.50 & 0.76 & 0.85 & 0.52 & 0.46 \\
LoReFT
& \textbf{0.80} & \textbf{0.89} & 0.77 & 0.87 & 0.77 & 0.86 \\
\midrule
\method{}~(Ours)
& \textbf{0.80} & 0.88
& \textbf{0.82} & \textbf{0.90}
& \textbf{0.85} & \textbf{0.92} \\
\bottomrule
\end{tabular}
}
\end{table}

\subsection{Analysis: Multi-Constraint Editing}
\label{sec:multi_constraint_analysis}
\textbf{RQ3:} Does \method{} apply constraint-aligned edits to the token positions where each constraint should be realized?
\textbf{Finding 3: \method{} performs position-aware, constraint-aligned token editing.}
We use the \textit{start\_with} constraint as a diagnostic case to examine whether \method{} applies multi-constraint edits to the appropriate token positions.
Intuitively, if a joint textual condition contains a \textit{start\_with} requirement, then the corresponding activation change should be most relevant at the beginning of the response rather than being uniformly applied to all tokens.

To test this, we need a reference direction that represents the \textit{start\_with} concept in activation space.
Following AxBench~\citep{wu2025axbench}, which shows that CAA directions can serve as effective concept detectors, we use the CAA direction for each constraint type as a reference concept axis.
This indicates that \method{} does not simply apply a global perturbation to all token activations.
For a constraint type $r$, such as \textit{start\_with}, we compute a CAA direction $\mathbf{v}^{(\ell)}_r$ from positive and negative examples.
We then compare this reference direction with the token-level edit direction produced by \method{}:
\[
\Delta \mathbf{a}^{(\ell)}_i =
\hat{\mathbf{a}}^{(\ell)}_{\mathrm{edit},i}
-
\mathbf{a}^{(\ell)}_{\mathrm{src},i},
\]
and measure their cosine similarity,
\[
s^{(\ell)}_{i,r}
=
\cos
\left(
\Delta \mathbf{a}^{(\ell)}_i,
\mathbf{v}^{(\ell)}_r
\right).
\]
This score measures whether the edit at token position $i$ moves the activation toward the direction associated with constraint $r$.

\begin{figure}[t]
    \centering
    \includegraphics[width=\columnwidth]{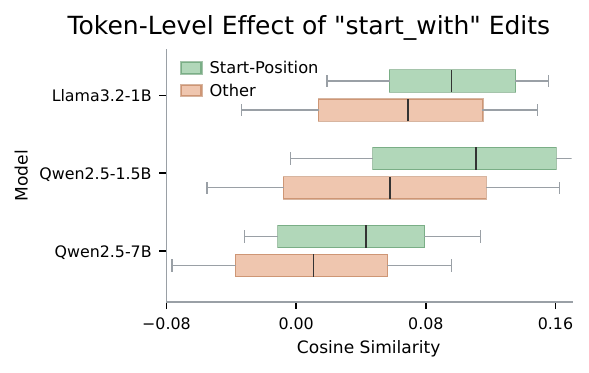}
    \caption{
    Token-level alignment between \method{} edits and CAA constraint directions.
    For the \textit{start\_with} constraint, edits at start-position tokens show higher cosine similarity with the CAA \textit{start\_with} direction than edits at other positions.
    }
    \label{fig:token_constraint_alignment}
\end{figure}

Figure~\ref{fig:token_constraint_alignment} shows a clear position-sensitive pattern.
For the \textit{start\_with} constraint, edit directions at start-position tokens are substantially more aligned with the CAA \textit{start\_with} direction than edits at middle or ending positions.
This indicates that \method{} does not simply apply a global perturbation to all token activations.
Instead, it produces stronger constraint-aligned updates at the positions where the constraint should be realized.

This analysis provides token-level evidence for how \method{} handles multi-constraint editing.
A joint condition may contain requirements that affect different parts of the response, such as beginning tokens, ending tokens, formatting tokens, or semantic content tokens.
The observed alignment pattern suggests that \method{} can translate a compositional textual condition into localized activation updates, where different token positions are edited toward the concept directions relevant to the constraints they need to satisfy.



\section{Conclusion}
\label{sec:conclusion}
We presented \method{}, a text-conditioned activation flow model that provides a unified interface for LLM steering and activation-space classification.
By learning a unified conditional velocity field over residual-stream activations, \method{} edits activations through flow inversion without fitting separate directions or intervention modules for each target behavior.
Experiments across three target LLMs show that \method{} performs strongly across behavioral control, TruthfulQA Truth*Info steering, fine-grained concept steering, multi-constraint instruction following, and activation-space classification.

\section{Limitations and Safety Discussion}

Although our evaluations cover behavioral control, truthfulness steering, concept steering, multi-constraint instruction following, and activation-space classification, they do not fully characterize \method{}'s effects on broader model capabilities.
We have not yet evaluated long-form generation, multi-turn stability, or complex reasoning tasks such as multi-step mathematics and planning.

\method{} also introduces safety considerations.
Because it can steer model behavior through natural-language conditions, the same mechanism that improves helpfulness, truthfulness, or constraint satisfaction could in principle be used to amplify undesirable behaviors, such as sycophancy, deception, or harmful personas.
In this work, conditions such as harmful or adversarial personas are used only as controlled evaluation targets for measuring activation-level controllability.
Future releases of trained activation-flow models should consider restricting unsafe target conditions, adding condition-level safety filters, and auditing edited generations with external safety classifiers or human review.




\bibliography{custom}

\newpage
\appendix

\section{Details of Conditional Flow Matching}
\label{app:flow_details}

This section provides additional details for the conditional flow-matching objective in Eq.~\ref{eq:flow_matching}.
Throughout this section, we use 
$\mathbf{a}$ as shorthand for the residual-stream activation 
$\mathbf{a}^{(\ell)}_i$ at layer $\ell$ and token position $i$.

\subsection{Interpolation Path and Target Velocity}
\label{app:flow_path}

For each training example, we obtain an activation-condition tuple
$(\mathbf{a}^{(\ell)}_i,\mathbf{c},\ell,i)$ from the frozen target language model $\mathcal{M}$.
We sample a prior activation state
\begin{equation}
    \mathbf{a}_0 \sim \mathcal{N}(\mathbf{0}, \mathbf{I}),
\end{equation}
with the same dimensionality as $\mathbf{a}^{(\ell)}_i$.
For readability, let $\mathbf{a}_1=\mathbf{a}^{(\ell)}_i$ denote the target activation.
We use a linear interpolation path:
\begin{equation}
\label{eq:app_interpolation}
    \mathbf{a}_t = (1-t)\mathbf{a}_0 + t\mathbf{a}_1,
    \qquad t \sim \mathcal{U}(0,1).
\end{equation}
The corresponding target velocity is
\begin{equation}
\label{eq:app_velocity}
    \mathbf{u}_t
    =
    \frac{\dd \mathbf{a}_t}{\dd t}
    =
    \mathbf{a}_1 - \mathbf{a}_0.
\end{equation}
This provides the velocity target $\mathbf{u}_t$ used in Eq.~\ref{eq:flow_matching}.
Although $\mathbf{u}_t$ is constant along this linear path, the model still receives $t$ as input and learns a time-dependent velocity field.

\subsection{Conditional Flow-Matching Objective}
\label{app:cfm_objective}

Following the notation in the main text, the velocity network is conditioned on the interpolated activation $\mathbf{a}_t$, timestep $t$, textual condition $\mathbf{c}$, layer index $\ell$, and token position $i$.
Substituting the linear-path velocity from Eq.~\ref{eq:app_velocity} into Eq.~\ref{eq:flow_matching}, the training objective becomes
\begin{equation}
\label{eq:app_fm_loss_expanded}
\begin{aligned}
    \mathcal{L}_{\mathrm{FM}}
    =
    \mathbb{E}_{(\mathbf{a}^{(\ell)}_i,\mathbf{c},\ell,i),\,\mathbf{a}_0,\,t}
    \Big[
    \big\|
    &v_\theta(\mathbf{a}_t,t,\mathbf{c},\ell,i) \\
    &-
    (\mathbf{a}^{(\ell)}_i-\mathbf{a}_0)
    \big\|_2^2
    \Big].
\end{aligned}
\end{equation}
This is the expanded form of Eq.~\ref{eq:flow_matching}.
The learned vector field induces the conditional ODE in Eq.~\ref{eq:activation_ode}:
\begin{equation}
\label{eq:app_activation_ode}
    \frac{\dd \mathbf{a}_t}{\dd t}
    =
    v_\theta(\mathbf{a}_t,t,\mathbf{c},\ell,i),
    \qquad t\in[0,1].
\end{equation}
Solving this ODE from $t=0$ to $t=1$ maps a prior activation state $\mathbf{a}_0$ to a condition-specific activation, while solving it backward maps an observed activation toward the prior.

\subsection{Conditional Flow Map}
\label{app:flow_map}

As in the main text, we denote by
$F_{\theta}^{s\rightarrow t}(\cdot;\mathbf{c},\ell,i)$
the flow map induced by Eq.~\ref{eq:activation_ode}.
For an activation state $\mathbf{a}_s$ at time $s$, the transported state is
\begin{equation}
\label{eq:app_flow_map}
    \mathbf{a}_t
    =
    F_{\theta}^{s\rightarrow t}
    (\mathbf{a}_s;\mathbf{c},\ell,i).
\end{equation}
In practice, this map is computed by numerically integrating the learned ODE.
Forward integration from $0$ to $1$ performs condition-specific activation generation, while backward integration from $1$ to an intermediate timestep performs activation inversion.

\subsection{Flow-Inversion Editing}
\label{app:flow_inversion_details}

This subsection expands the editing operation in Eq.~\ref{eq:flow_inversion_editing}.
Given a source activation $\mathbf{a}_{\mathrm{src}}$, a source condition $\mathbf{c}_{\mathrm{src}}$, and a target condition $\mathbf{c}_{\mathrm{tgt}}$, \method{} first partially inverts the source activation under the source-conditioned flow:
\begin{equation}
\label{eq:app_source_inversion}
    \mathbf{a}_{\tau}
    =
    F_{\theta}^{1\rightarrow \tau}
    (\mathbf{a}_{\mathrm{src}};\mathbf{c}_{\mathrm{src}},\ell,i),
\end{equation}
where $\tau=1-\lambda$ and $\lambda\in[0,1]$ is the edit strength.
It then follows the target-conditioned flow forward:
\begin{equation}
\label{eq:app_target_regeneration}
    \hat{\mathbf{a}}_{\mathrm{edit}}
    =
    F_{\theta}^{\tau\rightarrow 1}
    (\mathbf{a}_{\tau};\mathbf{c}_{\mathrm{tgt}},\ell,i).
\end{equation}
Combining the two steps gives
where $\tau=1-\lambda$ denotes the inversion timestep. The full editing operation can be written as
\begin{equation}
\label{eq:app_full_edit}
\begin{aligned}
    \mathbf{a}_{\tau}
    &=
    F_{\theta}^{1\rightarrow \tau}
    (\mathbf{a}_{\mathrm{src}};\mathbf{c}_{\mathrm{src}},\ell,i), \\
    \hat{\mathbf{a}}_{\mathrm{edit}}
    &=
    F_{\theta}^{\tau\rightarrow 1}
    (\mathbf{a}_{\tau};\mathbf{c}_{\mathrm{tgt}},\ell,i).
\end{aligned}
\end{equation}

When $\lambda$ is small, $\tau$ is close to $1$, so the inversion is shallow and the edited activation remains close to $\mathbf{a}_{\mathrm{src}}$.
Conversely, when $\lambda$ is large, $\tau$ approaches $0$, resulting in a deeper inversion and a more pronounced editing effect, which drives the edited activation $\mathbf{a}_{\mathrm{edit}}$ significantly away from $\mathbf{a}_{\mathrm{src}}$.

\section{Training Corpus Construction}
\label{app:training_corpus}

This section describes how we construct the activation-condition corpus used to train \method{}.
The corpus consists of tuples
$(\mathbf{a}^{(\ell)}_i,\mathbf{c},\ell,i)$,
where $\mathbf{a}^{(\ell)}_i$ is a residual-stream activation extracted from a frozen target language model $\mathcal{M}$ at layer $\ell$ and token position $i$, and $\mathbf{c}$ is a textual condition describing the behavior, concept, constraint, or label associated with the activation.
The same construction procedure is applied independently for each target LLM.

\subsection{Activation Extraction}
\label{app:activation_extraction}

Given a text sequence $\mathbf{x}=(x_1,\ldots,x_n)$, we run the frozen target language model $\mathcal{M}$ with teacher forcing and extract residual-stream activations from selected layers and token positions.
For layer $\ell$ and token position $i$, the extracted activation is denoted as
\begin{equation}
    \mathcal{A}(\mathbf{x})
    =
    \left\{
    \mathbf{a}^{(\ell)}_i
    :
    \ell \in \mathcal{L},
    i \in \mathcal{I}(\mathbf{x})
    \right\}.
\end{equation}
We use token positions from the response portion for all training data.


\subsection{Training Data Mixture}
\label{app:data_mixture}
The training corpus combines heterogeneous supervision sources and converts them into the same activation-condition format.
Each training example provides a text sequence, a supervision signal, and a textual condition $\mathbf{c}$ derived from that signal.
After running the frozen target model $\mathcal{M}$, the extracted activations are paired with $\mathbf{c}$ to form tuples
$(\mathbf{a}^{(\ell)}_i,\mathbf{c},\ell,i)$.

We group the training sources into three broad categories.
First, \textbf{behavioral supervision} contains examples associated with high-level generation behaviors, such as persona traits, truthfulness, helpfulness, harmlessness, refusal, sycophancy, and hallucination.
Behavioral supervision is constructed from Persona Vectors, HH-RLHF, and HelpSteer, while TruthfulQA is used only for evaluation.

Second, \textbf{fine-grained concept supervision} contains examples associated with localized semantic concepts, where the textual condition describes the target concept.
We use AxBench Concept500 as the main source for concept-conditioned activation modeling, where each concept description is converted into a textual condition.

Third, \textbf{constraint-following supervision} contains examples with explicit output requirements, including multi-constraint settings.
We use the training part of RECAST-5 and RECAST-10 to provide compositional constraint-following examples, where multiple constraints are verbalized into a single textual condition.

All supervision signals are verbalized as natural-language conditions.
When a dataset provides categorical labels, scalar attributes, or rule annotations, we convert them with short templates.
For example, behavioral labels are verbalized as conditions such as
\textit{``Be evil.''}.

For compositional settings, multiple requirements are merged into a single condition string rather than represented as separate steering components.
For example, a multi-constraint condition may be written as
\textit{``The response should be concise, harmless, and end with the specified phrase.''}

\section{Implementation Details}
\label{app:implementation}

\subsection{Architecture}
\label{app:architecture}
\method{} is implemented as a DiT-style\citep{peebles2023scalablediffusionmodelstransformers} text-conditioned flow model over residual-stream activations.
For each target LLM, the model predicts the velocity field
$v_\theta(\mathbf{a}_t,t,\mathbf{c},\ell,i)$
defined in Eq.~\ref{eq:flow_matching}.
The input activation state $\mathbf{a}_t$ has the same dimensionality as the residual-stream activation $\mathbf{a}^{(\ell)}_i$ of the corresponding target LLM, and the output velocity has the same dimension.

We use Qwen3-Embedding-0.6B as the condition encoder.
Given a textual condition $\mathbf{c}$, the encoder produces a condition representation
\begin{equation}
    \mathbf{e}_{c}
    =
    \mathrm{Enc}(\mathbf{c}),
\end{equation}
where the parameters of $ \mathrm{Enc}$ are frozen during training.
The condition representation is then projected to the hidden dimension of the activation flow model:
\begin{equation}
    \tilde{\mathbf{e}}_{c}
    =
    W_c \mathbf{e}_{c} + \mathbf{b}_c .
\end{equation}

\subsection{Classifier-Free Guidance}
\label{app:cfg}

We train \method{} with classifier-free guidance~\citep{ho2022classifierfreediffusionguidance} to improve conditional controllability at inference time.
During training, the textual condition $\mathbf{c}$ is randomly replaced with a null condition $\varnothing$ with probability $p_{\mathrm{drop}}$.
The model is therefore trained to predict both conditional and unconditional velocity fields:    $v_\theta(\mathbf{a}_t,t,\mathbf{c},\ell,i)$ and $v_\theta(\mathbf{a}_t,t,\varnothing,\ell,i).$
The same flow-matching loss in Eq.~\ref{eq:flow_matching} is used for both conditional and unconditional training examples.


\subsection{Optimization}
\label{app:optimization}

For each target LLM, all parameters of the target model $\mathcal{M}$ are frozen.
The Qwen3-Embedding-0.6B condition encoder is also frozen.
Only the DiT-based activation flow model is trained.

For each batch, we sample activation-condition tuples
$(\mathbf{a}^{(\ell)}_i,\mathbf{c},\ell,i)$ from the training corpus $\mathcal{D}$.
We sample an activation state
$\mathbf{a}_0\sim\mathcal{N}(\mathbf{0},\mathbf{I})$
and a timestep $t\sim\mathcal{U}(0,1)$.
The interpolated activation $\mathbf{a}_t$ and target velocity $\mathbf{u}_t$ are computed as described in Appendix~\ref{app:flow_path}.
With probability $p_{\mathrm{drop}}$, the textual condition is replaced by the null condition $\varnothing$ for classifier-free guidance training.
The model is optimized using the flow-matching objective in Eq.~\ref{eq:flow_matching}.

We use AdamW with apeak learning rate of $4\times 10^{-5}$ and a cosine learning-rate schedule with linear warmup. All models are trained for $10$ epochs on approximately $270$K training examples. Training is performed on two GPUs with
gradient accumulation of $8$ steps. The per-GPU batch size is $2$ for Llama-3.2-1B and $4$ for both Qwen2.5-1.5B
and Qwen2.5-7B.


\section{Experimental Details}
\label{app:eval}
This section contains specific experiment details.
\subsection{Base models}
\label{app:base_models}
All of the language models are listed as follows:
\begin{itemize}
    \item For Llama-3.2-1B-Instruct model, we use \texttt{meta-llama/Llama-3.2-1B-Instruct}\footnote{https://huggingface.co/meta-llama/Llama-3.2-1B-Instruct}

    \item For Qwen2.5-1.5B-Instruct model, we use \texttt{Qwen/Qwen2.5-1.5B-Instruct}\footnote{https://huggingface.co/Qwen/Qwen2.5-1.5B-Instruct}

    \item  For Qwen2.5-7B-Instruct model, we use \texttt{Qwen/Qwen2.5-7B-Instruct}\footnote{https://huggingface.co/Qwen/Qwen2.5-7B-Instruct}
\end{itemize}

\subsection{Benchmarks and Metrics}
\label{app:benchmarks}
To ensure a fair evaluation, all activation steering methods are constructed from the same data mixture in Appendix~\ref{app:data_mixture}. Methods that learn trainable steering modules, including \method{}, use this mixture as their training data; methods that estimate explicit steering directions, such as CAA, use the same mixture for direction extraction.
The data mixture is designed to cover all attribute families that may appear in the evaluation benchmarks, and therefore benchmarks are merely for testing.
\paragraph{Persona} We test three traits on persona vectors dataset: evil, hallucinating, sycophantic. There are 20 questions for each trait, and each question is sampled 10 times. To judge, two scores are used: target-trait score to show how obvious the target-trait is represented and coherence score to show whether model's expressiveness is influenced. Both scores range from 0 to 100 and we calculate the average target-trait score over generations whose coherence score is above 40 as the final result.

\paragraph{TruthfulQA}
We evaluate whether steering methods can improve truthful answering. 
For each method, we generate responses to 817 TruthfulQA questions and evaluate them using the \texttt{allenai/truthfulqa-truth-judge-llama2-7B} judge model.
The judge provides truthfulness and informativeness scores for each response.
We report Truth*Info, computed as the percentage of responses that are judged both truthful and informative.

\paragraph{AxBench}
We use Concept10 evaluation from AxBench Dataset, for each concept, we randomly sample 10 instructions from alpaca\_eval, and make model generate response related with corresponding concept. Then we use LLM-as-judge to generate three scores: concept relevance, instruction relevance and fluency, all ranging from [0,1,2]. Then we use Harmonic Mean of them as final score.
\paragraph{RECAST}
We use RECAST-5 and RECAST-10 evaluation from RECAST Dataset, which separately have up to 5 and 10 constraints for one instruction. Inputs are instructions with constraints and we use Rule-based Constraint Satisfaction Rate (RSR) which requires a response obey all rule-based constraints.
\paragraph{ToxiGen}
We use ToxiGen to evaluate whether activation-space scoring can serve as a binary classifier for toxic versus non-toxic content.
For baselines, we follow the estimation protocol of \citet{wu2025axbench}.
For \method{}, we define two candidate textual conditions corresponding to toxic and non-toxic content, compute the conditional reconstruction energy under each condition, and predict the label with the lower energy.
We report both accuracy and ROC-AUC.
\subsection{Baseline Implementation}
\label{app:baseline_details}
We briefly describe each activation-steering baseline used in our comparison. For fairness, all baselines use the shared data mixture described in Appendix~\ref{app:data_mixture}: learning-based methods train their intervention modules on this mixture, while direction-based methods extract their steering directions from the same mixture.
\begin{itemize}
  \item \textbf{Contrastive Activation Addition (CAA)} computes the mean difference between positive and negative activations and uses this average difference as a fixed steering direction at inference time \citep{panickssery2023steering}. Scalar steering coefficient $\alpha$  is used to control the intervention strength. 
  \item \textbf{ODESteer} formulates activation steering as a continuous ODE-based editing process. Instead of applying a single additive shift, it integrates a steering dynamics over multiple steps and uses the resulting trajectory to modify hidden activations \citep{zhao2026odesteer}. The number of integration steps $T$ is introduced to control the intervention granularity. 
  We perform a grid search over $T$ on the validation split and report test results using the value that achieves the best validation performance.
  \item \textbf{LoReFT} parameterizes representation interventions with a low-rank transformation and learns a small set of intervention parameters while keeping the base language model frozen \citep{wu2024reftrepresentationfinetuninglanguage}.  we introduce a scalar steering coefficient $\alpha$ to control the
  \[
   \mathbf{a}' = \mathbf{a} + \alpha \mathbf{R}^T (\mathbf{W}\mathbf{a} + \mathbf{b} - \mathbf{R}\mathbf{a})
   \]
   where $\mathbf{R},\mathbf{W},\mathbf{b}$ is learned in the training process.
  We perform a grid search over $\alpha$ on the validation split and report test results using the value that achieves the best validation performance.
  \item \textbf{Representation Engineering (RepE)} constructs contrastive representations for the target behavior and extracts a principal steering direction, typically using PCA over activation differences, which is then added to hidden states during generation \citep{zou2025representationengineeringtopdownapproach}. Scalar steering coefficient $\alpha$  is used to control the intervention strength. We perform a grid search over $\alpha$ on the validation split and report test results using the value that achieves the best validation performance.
\end{itemize}
\subsection{Generation and Decoding Settings}
\label{app:generation_settings}

  Table~\ref{tab:generation_model_settings} summarizes the model-level generation and intervention settings.
  For all target LLMs, we use the \textbf{default} template provided by the corresponding model family.
  For activation editing, \method{} injects the edited activation at the middle transformer layer of each target
  model.

  \begin{table}[t]
  \centering
  \caption{
  Model-level generation and intervention settings. 
  For RECAST-5 and RECAST-10, we use 512 as max tokens settings.
  }
  \label{tab:generation_model_settings}
  \small
  \setlength{\tabcolsep}{4pt}
  \resizebox{0.8\columnwidth}{!}{
  \begin{tabular}{lcc}
  \toprule
  \textbf{Target model} & \textbf{Max tokens} & \textbf{Injection layer(Index)} \\
  \midrule
  Llama-3.2-1B & 256 & 7 \\
  Qwen2.5-1.5B & 256 & 14 \\
  Qwen2.5-7B & 256 & 14 \\
  \bottomrule
  \end{tabular}
  }
  \end{table}

  Table~\ref{tab:generation_temperature_settings} reports the temperature used for each generation benchmark.

  \begin{table}[t]
  \centering
  \caption{
  Benchmark-level decoding temperature.
  }
  \label{tab:generation_temperature_settings}
  \small
  \setlength{\tabcolsep}{6pt}
  \begin{tabular}{lc}
  \toprule
  \textbf{Benchmark} & \textbf{Temperature} \\
  \midrule
  Persona & 1.0 \\
  TruthfulQA & 0.0 \\
  AxBench & 0.0 \\
  RECAST-5 & 0.0 \\
  RECAST-10 & 0.0 \\
  \bottomrule
  \end{tabular}
  \end{table}

  For \method{}, 
  all ODE integrations use the Euler solver.
  The number of integration steps, classifier-free guidance scale, and $\tau$ are tuned separately
  for each target model and benchmark, as shown in Table~\ref{tab:unisteer_generation_settings}.

\begin{table*}[t!]
\centering
\caption{
\method{} generation-time ODE editing settings.
All runs use the \textbf{Euler} solver.
Here $w$ denotes the CFG scale, $\tau=1-\lambda$ denotes the inversion timestep in
$F_{\theta}^{1\rightarrow\tau}$, and ``Inv. steps'' denotes the number of Euler steps used for
$F_{\theta}^{1\rightarrow\tau}$.
For each benchmark, we search $w$ over the specified interval with the listed step size.
}
\label{tab:unisteer_generation_settings}
\small
\setlength{\tabcolsep}{5pt}
\renewcommand{\arraystretch}{1.10}
\begin{tabular}{llcccc}
\toprule
\textbf{Benchmark}
& \textbf{Target model}
& \textbf{ODE steps}
& \textbf{Inv. steps}
& $\boldsymbol{\tau}$
& \textbf{CFG scale $w$} \\
\midrule

\multirow{3}{*}{Persona}
& Llama-3.2-1B
& 30 & 15 & 0.5 & $[8,29]$, step 3 \\
& Qwen2.5-1.5B
& 30 & 15 & 0.5 & $[8,30]$, step 3 \\
& Qwen2.5-7B
& 30 & 15 & 0.5 & $[8,30]$, step 3 \\

\midrule

\multirow{3}{*}{TruthfulQA}
& Llama-3.2-1B
& 20 & 10 & 0.5 & $[5,25]$, step 5 \\
& Qwen2.5-1.5B
& 20 & 10 & 0.5 & $[5,20]$, step 5 \\
& Qwen2.5-7B
& 20 & 10 & 0.5 & $\{7,10,13\}$ \\

\midrule

\multirow{3}{*}{AxBench}
& Llama-3.2-1B
& 50 & 30 & 0.4 & $[50,70]$, step 5 \\
& Qwen2.5-1.5B
& 50 & 30 & 0.4 & $[50,70]$, step 5 \\
& Qwen2.5-7B
& 50 & 30 & 0.4 & $[50,70]$, step 5 \\

\midrule

\multirow{3}{*}{RECAST-5}
& Llama-3.2-1B
& 10 & 1 & 0.9 & $[5,30]$, step 5 \\
& Qwen2.5-1.5B
& 10 & 1 & 0.9 & $[5,30]$, step 5 \\
& Qwen2.5-7B
& 10 & 1 & 0.9 & $[5,30]$, step 5 \\

\midrule

\multirow{3}{*}{RECAST-10}
& Llama-3.2-1B
& 10 & 1 & 0.9 & $[5,30]$, step 5 \\
& Qwen2.5-1.5B
& 10 & 1 & 0.9 & $[5,30]$, step 5 \\
& Qwen2.5-7B
& 10 & 1 & 0.9 & $[5,30]$, step 5 \\

\bottomrule
\end{tabular}
\end{table*}

\section{Additional Results}
\label{app:additional_results}

\subsection{Effect of Edit Strength}
\label{app:edit_strength}

\begin{figure*}[b] 
\centering

\begin{subfigure}[t]{0.46\textwidth}
    \centering
    \includegraphics[width=\linewidth]{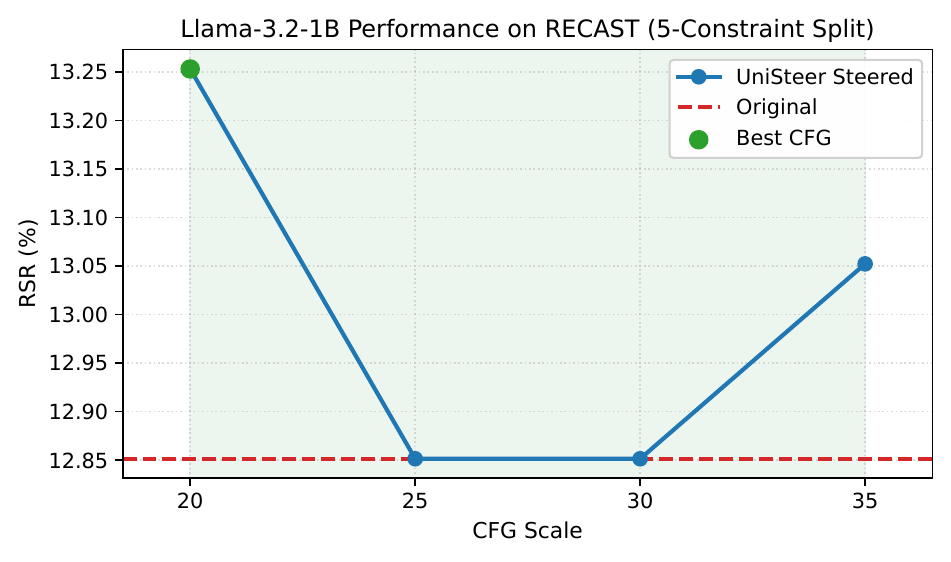}
    \caption{RECAST-5 sweep for Llama-3.2-1B.}
    \label{fig:recast_llama32_1b_r5}
\end{subfigure}
\hfill
\begin{subfigure}[t]{0.46\textwidth}
    \centering
    \includegraphics[width=\linewidth]{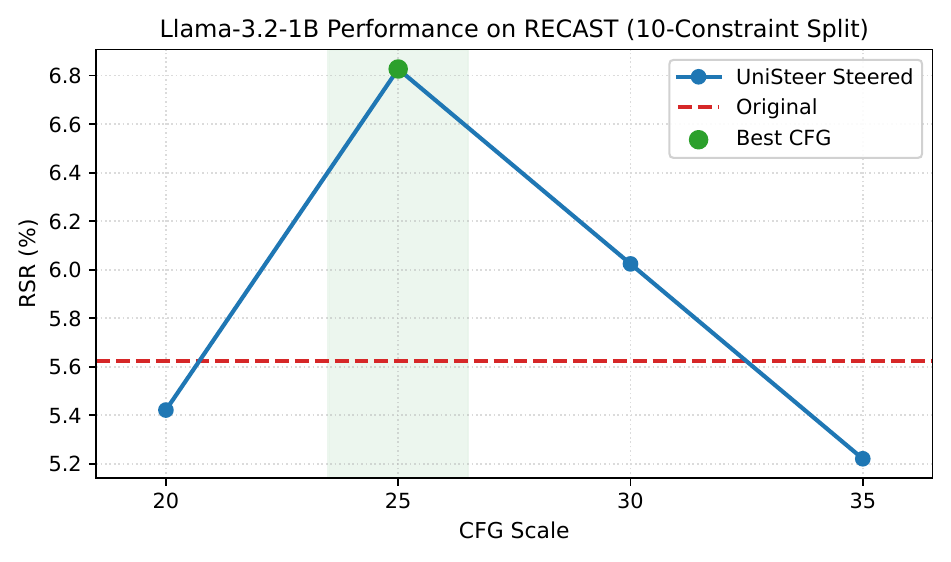}
    \caption{RECAST-10 sweep for Llama-3.2-1B.}
    \label{fig:recast_llama32_1b_r10}
\end{subfigure}
\vfil 

\begin{subfigure}[t]{0.46\textwidth}
    \centering
    \includegraphics[width=\linewidth]{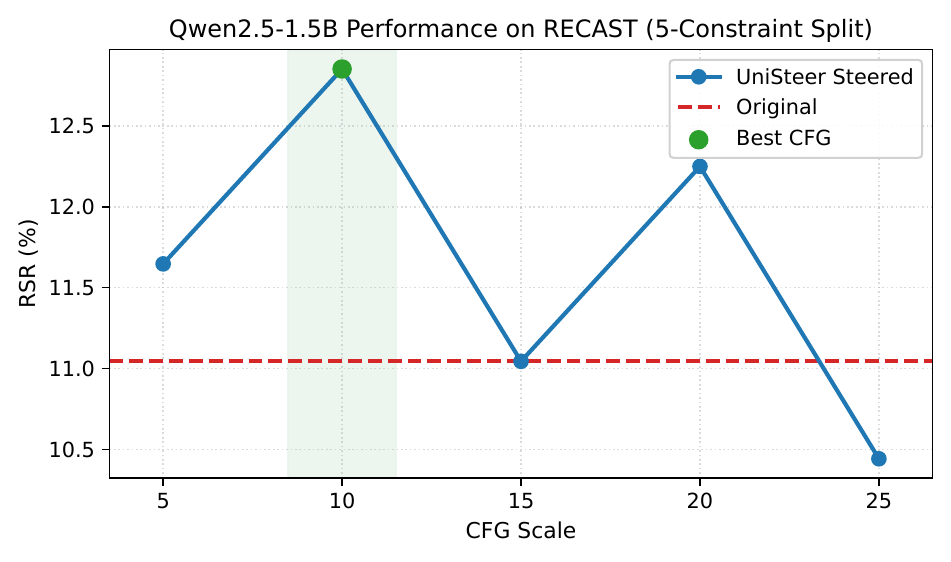}
    \caption{RECAST-5 sweep for Qwen2.5-1.5B.}
    \label{fig:recast_qwen25_15b_r5}
\end{subfigure}
\hfill
\begin{subfigure}[t]{0.46\textwidth}
    \centering
    \includegraphics[width=\linewidth]{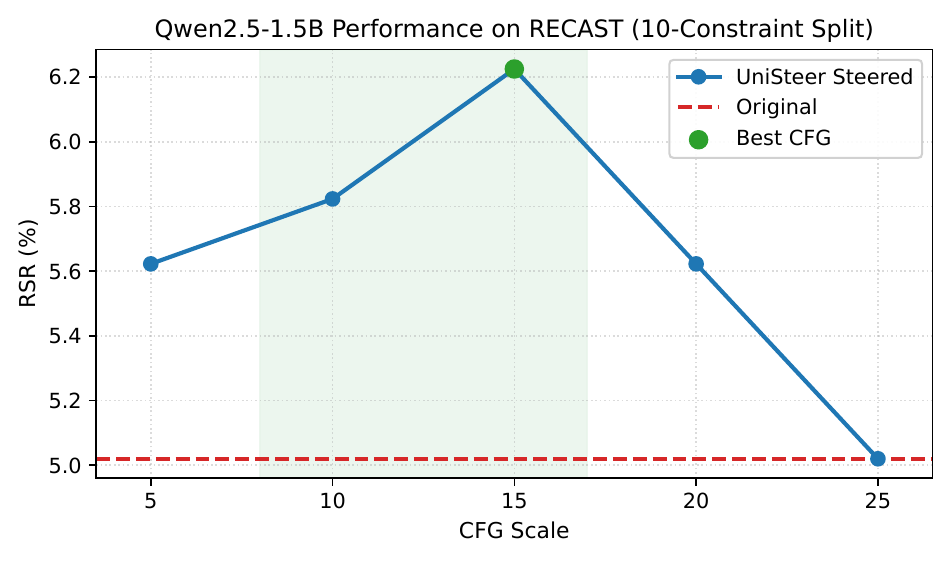}
    \caption{RECAST-10 sweep for Qwen2.5-1.5B.}
    \label{fig:recast_qwen25_15b_r10}
\end{subfigure}
\vfil

\begin{subfigure}[t]{0.46\textwidth}
    \centering
    \includegraphics[width=\linewidth]{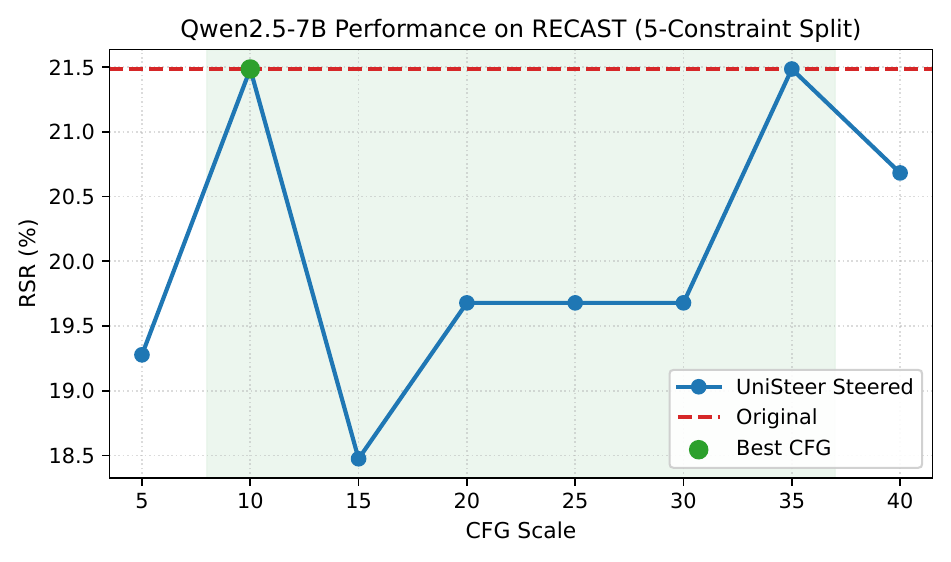}
    \caption{RECAST-5 sweep for Qwen2.5-7B.}
    \label{fig:recast_qwen25_7b_r5}
\end{subfigure}
\hfill
\begin{subfigure}[t]{0.46\textwidth}
    \centering
    \includegraphics[width=\linewidth]{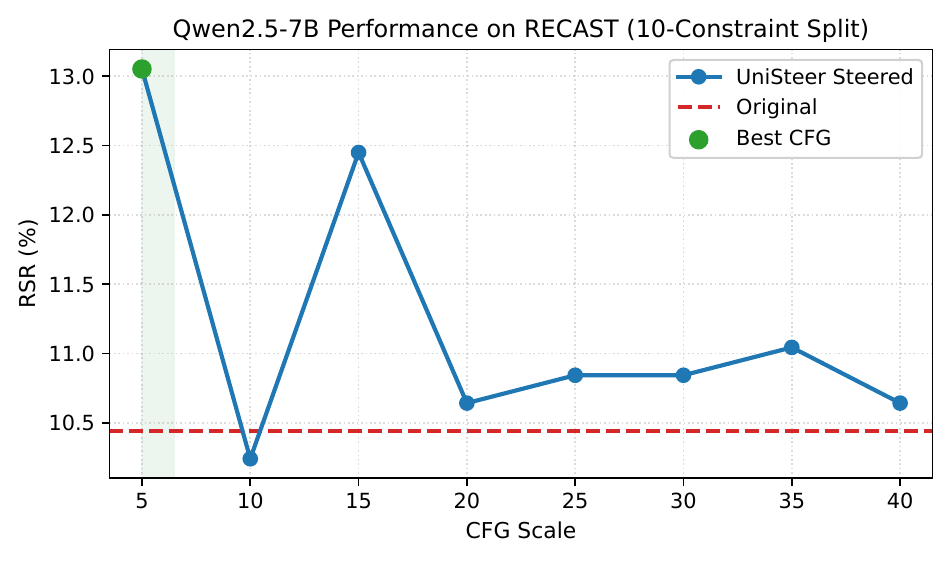}
    \caption{RECAST-10 sweep for Qwen2.5-7B.}
    \label{fig:recast_qwen25_7b_r10}
\end{subfigure}
\vfil

\begin{subfigure}[t]{0.46\textwidth}
    \centering
    \includegraphics[width=\linewidth]{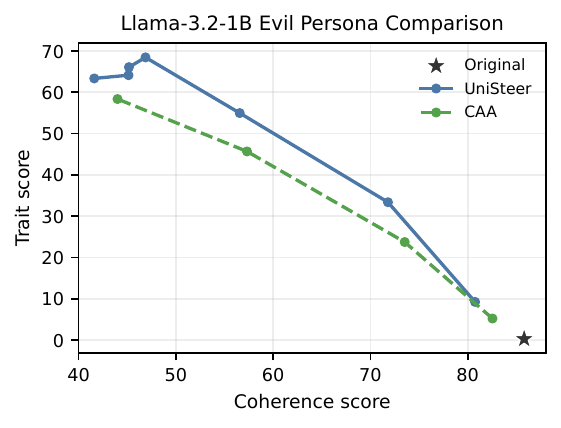}
    \caption{Llama-3.2-1B (Trait-coherence).}
    \label{fig:persona_trait_coherence_llama32_1b}
\end{subfigure}
\hfill
\begin{subfigure}[t]{0.46\textwidth}
    \centering
    \includegraphics[width=\linewidth]{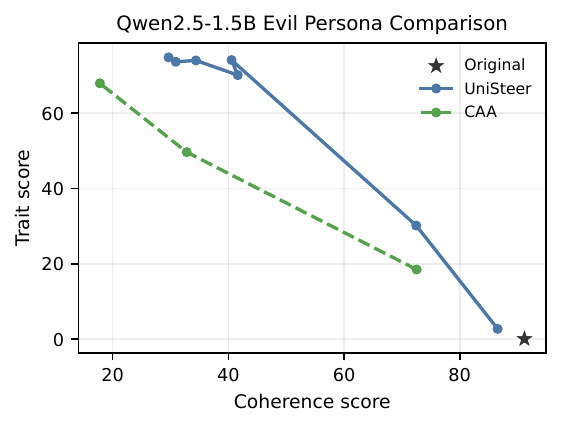}
    \caption{Qwen2.5-1.5B (Trait-coherence).}
    \label{fig:persona_trait_coherence_qwen25_15b}
\end{subfigure}

\caption{Hyperparameter sweeps on RECAST (top three rows) and Trait--coherence trade-off on the Persona evil trait (bottom row).}
\label{fig:combined_all_plots}
\end{figure*}

\paragraph{RECAST Hyperparameter Sensitivity.}
We further analyze the sensitivity of \method{} to the classifier-free guidance scale on RECAST.
In this analysis, we use 10 integration steps for flow inversion and reconstruct the activation after rolling back one step.
As shown in figures~\ref{fig:combined_all_plots}\subref{fig:recast_llama32_1b_r5}--\subref{fig:recast_qwen25_7b_r10}, the hyperparameter sweeps results for RECAST-5 and RECAST-10 across the three target LLMs.
Overall, the optimal guidance scale is both model-dependent and constraint-dependent.
For Llama-3.2-1B, a relatively large guidance scale works best on RECAST-5, while a moderate scale gives the best result on RECAST-10.
For Qwen2.5-1.5B, the best scale also shifts with the number of constraints, with RECAST-5 preferring a smaller-to-moderate value and RECAST-10 preferring a slightly stronger value.
For Qwen2.5-7B, \method{} improves over the original model across several CFG values and reaches the best result at a small CFG scale on RECAST-10.
However, on RECAST-5, the method does not consistently improve over the original model and remains below or close to the original RSR across the tested CFG range.
It indicates that activation editing can introduce unnecessary perturbations when the original model already follows the constraints, especially under aggressive guidance.

\end{document}